\documentclass[runningheads,a4paper]{llncs}

\usepackage[english]{babel}

\usepackage[%
rm={oldstyle=false,proportional=true},%
sf={oldstyle=false,proportional=true},%
tt={oldstyle=false,proportional=true,variable=true},%
qt=false%
]{cfr-lm}
\usepackage{graphicx}

\usepackage{paralist}

\usepackage{cite}

\usepackage[T1]{fontenc}

\usepackage[math]{blindtext}
\usepackage{amsmath}
\usepackage{amssymb}
\DeclareMathOperator*{\argmin}{argmin}
\newcommand*{\argminl}{\argmin\limits}

\usepackage{csquotes}

\usepackage{microtype}

\usepackage{booktabs}

\usepackage{url}
\makeatletter
\g@addto@macro{\UrlBreaks}{\UrlOrds}
\makeatother

\usepackage{xcolor}

\usepackage[
bookmarks=false,
breaklinks=true,
colorlinks=true,
linkcolor=black,
citecolor=black,
urlcolor=black,
pdfpagelayout=SinglePage,
pdfstartview=Fit
]{hyperref}
\usepackage[all]{hypcap}

\usepackage{pdfcomment}

\usepackage{xspace}

\DeclareFontFamily{U}{MnSymbolC}{}
\DeclareSymbolFont{MnSyC}{U}{MnSymbolC}{m}{n}
\DeclareFontShape{U}{MnSymbolC}{m}{n}{
    <-6>  MnSymbolC5
   <6-7>  MnSymbolC6
   <7-8>  MnSymbolC7
   <8-9>  MnSymbolC8
   <9-10> MnSymbolC9
  <10-12> MnSymbolC10
  <12->   MnSymbolC12%
}{}
\DeclareMathSymbol{\powerset}{\mathord}{MnSyC}{180}

\begin{document}

\input glyphtounicode.tex
\pdfgentounicode=1

\title{Motion Estimated-Compensated Reconstruction with Preserved-Features in Free-Breathing Cardiac MRI}
\titlerunning{Free-Breathing Reconstruction in Cardiac MRI}

\author{Firstname Lastname \and Firstname Lastname}
\institute{Institute}

%
\author{Aur\'elien Bustin\inst{1,2,3}, Anne Menini\inst{1}, Martin A. Janich\inst{1}, Darius Burschka\inst{2}, Jacques Felblinger\inst{3,4,5}, Anja C.S. Brau\inst{6}, \and Freddy Odille\inst{3,4,5} }
\authorrunning{A. Bustin et al.}

\institute{
GE Global Research Center, Munich, Germany\\
\and
Computer Science, Technische Universit\"at M\"unchen, Munich, Germany\\
\and
U947, INSERM, Nancy, France\\
\and
IADI, Universit\'e de Lorraine, Nancy, France\\
\and
CIC-IT 1433, INSERM, Nancy, France\\
\and
GE Healthcare, MR Applications and Workflow, Menlo Park, California, USA
}
			
\maketitle

\begin{abstract}
To develop an efficient motion-compensated reconstruction technique for free-breathing cardiac magnetic resonance imaging (MRI) that allows high-quality images to be reconstructed from multiple undersampled single-shot acquisitions. The proposed method is a joint image reconstruction and motion correction meth-od consisting of several steps, including a non-rigid motion extraction and a motion-compensated reconstruction. The reconstruction includes a denoising with the Beltrami regularization, which offers an ideal compromise between feature preservation and staircasing reduction. Results were assessed in simulation, phantom and volunteer experiments. The proposed joint image reconstruction and motion correction method exhibits visible quality improvement over previous methods while reconstructing sharper edges. Moreover, when the acceleration factor increases, standard methods show blurry results while the proposed meth-od preserves image quality. The method was applied to free-breathing single-shot cardiac MRI, successfully achieving high image quality and higher spatial resolu-tion than conventional segmented methods, with the potential to offer high-quality delayed enhancement scans in challenging patients.
\end{abstract}


\section{Introduction}\label{sec:intro}
Cardiac magnetic resonance imaging is a valuable tool for myocardial structure, function, and tissue assessment, providing essential information for clinical diagnosis and treatment decisions in cardiovascular disease. Using standard segmented sequences in which data acquisition is segmented over multiple heart beats, good image quality can be obtained in patients with regular cardiac rhythm and good breath-holding ability; however, image quality can be degraded by motion artifacts when scanning patients with arrhythmia or poor breath-hold compliance. In comparison to segmented acquisitions, single-shot techniques can be applied for rapid image acquisition of a whole slice within a single shot, greatly reducing the scan time. Due to the short acquisition duration of single-shot techniques (typically < 200 ms), artifacts from intra-shot motion are negligible, therefore such methods tend to be robust against cardiac and breathing motion. However, this motion robustness comes at the expense of lower spatial resolution and signal to noise ratio (SNR). An example of the benefit of single-shot over segmented Late Gadolinium Enhanced (LGE) imaging in a patient who could not breath-hold is shown in \ref{fig:fig1}. Recent techniques proposed to enhance the SNR of single-shot methods by motion correcting and then averaging multiple single-shot images acquired in free-breathing \cite{Kellman2005}. While this technique shows good results with low acceleration factors, it may not provide optimal image quality for higher undersampling, introducing blurring and undersampling artifacts, mainly due to higher weight given to the regularization.  Batchelor et al. \cite{Batchelor2005} proposed a first generalized reconstruction framework for motion compensation. The method allows arbitrary motion to be compensated by solving a general matrix inversion problem. This technique, however, requires an adequate knowledge of the displacement fields. The recent GRICS method \cite{Odille2008a} extended this work by jointly estimating the motion and the recovered image, however, it relied on a motion model provided by external sensors (e.g. ECG, respiratory belt).

In this work, we sought to develop an efficient motion correction implementation suitable for reconstructing a high-resolution, high-SNR image from multiple accelerated single-shot images. The proposed method combines the benefits of using a hybrid self-navigated sampling scheme (see Fig. \ref{fig:fig1}) with a joint reconstruction framework. In the image reconstruction step, a highly efficient feature-preserving regularization scheme (Beltrami) is proposed for recovering sharp details. We show that the proposed method is robust to high acceleration factors and yields results with efficient noise reduction and better overall image quality at a low computational cost.

\section{Theory}

\subsection{General Motion Compensation Framework}

Motion compensation techniques aim to solve the following inverse problem: find an underlying image $\rho$ free of motion artifacts, given derived measurements $s$ through the system $E$, affected by noise $\nu: s = E\rho + \nu$. Where $E$ is the encoding matrix, generally composed of a sampling operator $\xi$, a Fourier transform $F$, coil sensitivity maps $\sigma$ (in case of parallel imaging reconstruction), and a motion warping operator $W$ describing a non-rigid deformation for each shot. Here $\rho \in \mathbb{C}^{n_x \times n_y}$ and $s \in \mathbb{C}^{n_x \times n_y/acc \times n_r \times n_c}$  are complex data with $n_c$ the number of receiver coils, $n_r$ the number of repetitions and $acc$ the acceleration factor. In this work, the acquired data s represents the k-space data from multiple single-shot images and is generally corrupted by noise. A typical approach to solve this problem is to minimize the squared difference as assessed by the Euclidean norm. However, this problem is generally ill-posed (e.g. due to undersampling and noise), leading to non-uniqueness of the solution, if it exists. Thus, regularity constraints on the unknown solution $\rho$ have to be considered. Furthermore, motion should also be considered as unknown. The general joint optimization framework is then defined as

\begin{equation}
\label{eq:eq1}
\rho = \argminl_{\left( \rho, \vec{u}\right)} \lbrace \Vert s - E\left(\vec{u}\right) \rho \Vert_2^2 + \lambda \phi \left( \rho \right) \rbrace \text{ where } E\left( \vec{u}\right) = \xi F \sigma W \left( \vec{u}\right)
\end{equation}

Here $\vec{u}$ represents the displacement fields, $\phi$ is the chosen regularization function and $\lambda > 0$ is the corresponding regularization parameter. The optimization problem in Equation \ref{eq:eq1} is solved in four steps: (i) we first use the k-space center of the single-shot images to extract a self-navigation signal and to cluster the raw data into a reduced number of respiratory bins \cite{Usman2013}; (ii) we reconstruct the images from each bin independently using a Beltrami-regularized SENSE (B-SENSE) reconstruction; (iii) then an estimate of the motion is obtained using a non-rigid registration (minimization of Equation \ref{eq:eq1} with regards to $\vec{u}$ \cite{Odille2014}) and (iv) a high resolution/SNR image is generated using the proposed motion-compensated reconstruction process (minimization with regards to $\rho$). A general description of the method is shown in Fig. \ref{fig:fig2}.

\begin{figure}[h!]
\centering
\includegraphics[width=0.91\textwidth]{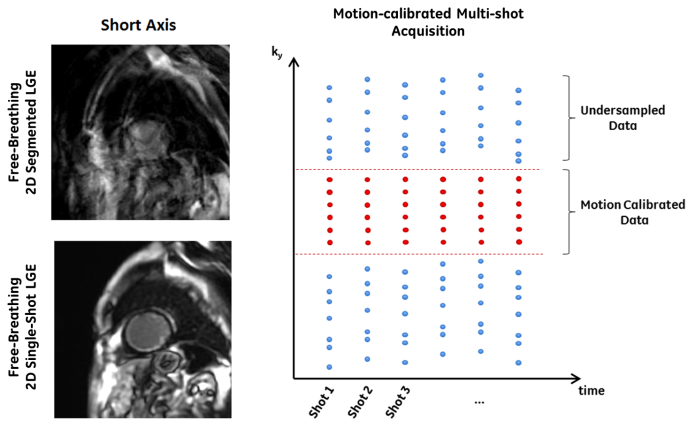}
\caption{(Left) Comparison between 2D segmented LGE (top) and 2D single-shot LGE (bottom) on a 77-year-old patient with breath-holding difficulties. The segmented LGE has higher resolution than the single-shot LGE but shows severe motion artefacts. (Right) Proposed hybrid k-space acquisition scheme including motion-calibration data (fully sampled center) and undersampled periphery, aimed to combine the resolution of segmented LGE with the motion-robustness of single-shot LGE.}
\label{fig:fig1}
\end{figure}

\subsection{Beltrami-Regularized SENSE}

In our framework, a respiratory signal is extracted from the motion calibration data itself. This pre-processing step is achieved by stacking the low resolution images along the time dimension. Singular value decomposition is then applied to the stack and the first left-singular vector is used as a good approximation of the true respiratory signal. A specific respiratory phase is then assigned to each acquired shot, as explained in \cite{Usman2013}. This binning strategy would split the data into fewer motion states $n_b \left( < n_r\right)$ with negligible respiration motion and lower undersampling in each of them. Images from each respiratory bin $\left( \rho_i \right)_{i = 1,\dots,n_b}$ are then individually reconstructed by solving

\begin{equation}
\label{eq:eq2}
\rho_i = \argminl_{\rho} \lbrace \Vert s_i - E_i \rho \Vert_2^2 + \lambda \sqrt{1+\beta^2 \vert \nabla \rho \vert^2} \rbrace
\end{equation}

The first term in Equation \ref{eq:eq2} is a data fidelity term that aims to minimize the difference between the reconstructed image and the acquired data. The Beltrami regularization $\sqrt{1+\beta^2 \vert \nabla \rho \vert^2}$ has been introduced in the field of string theory for physics and has shown high potential in several imaging problems, including image denoising and enhancement \cite{Polyakov1981} and super-resolution reconstruction \cite{Odille2015}. In particular, the metric can be chosen such that the Beltrami energy corresponds to an arbitrary interpolation between Gaussian diffusion $\beta \to 0$ and total variation (TV) \cite{Rudin1992a} regularization $\beta \gg 1$. In \cite{Zosso2014}, the authors showed that Beltrami regularization is able to maintain the advantage of TV (edges preserving, noise reduction) as well as reducing the effect of staircasing. B-SENSE is very similar to compressed sensing SENSE (CS-SENSE) methods presented by other authors \cite{Liang2008}, where here Beltrami is making the image sparse in the gradient domain. Even though this suggests that B-SENSE has a close relationship with the compressed sensing (CS) theory, it is, however, not CS as defined by Cand\`es et al. \cite{Candes2006}, especially due to the pseudo-random undersampling pattern used here (i.e. a uniform random pattern is used in \cite{Candes2006}). We propose to solve Equation \ref{eq:eq2} by adopting a primal-dual projected gradient approach \cite{Chan1999} with the potential to converge faster than the classic primal gradient-descent \cite{Zosso2014}. Respiratory motion estimation is then accomplished using independent non-rigid registration of the images reconstructed from each respiratory bin. Here we use an iterative framework validated in a large patient database for myocardial $T_2$ mapping \cite{Odille2014}, which is based on minimizing the sum-of-squared differences of the pixel intensities within a multi-resolution Gauss-Newton scheme.

\subsection{Motion Compensated Reconstruction with Preserved-Features}

This section presents the final step for solving the motion compensated problem in Equation \ref{eq:eq1}. The aim of the method is to reconstruct the high resolution, high SNR image $\rho$ from the acquired raw data $s = \left( s_i \right)_{i = 1,\dots,n_b}$. For the motion compensated reconstruction, we solve the following optimization problem, with the acquisition model now including the estimated motion fields:

\begin{equation}
\label{eq:eq3}
\rho = \argminl_{\rho} \lbrace \Vert s - E\left(\vec{u}\right) \rho \Vert_2^2 + \lambda \sqrt{1+\beta^2 \vert \nabla \rho \vert^2} \rbrace
\end{equation}

As in the previous section, we use a primal-dual projected gradient approach, employing the Beltrami energy as regularity prior \cite{Zosso2014}. Note that regularization is always preferred in motion compensated reconstruction due to the ill-conditioning induced by the motion operators, as shown in \cite{Atkinson2003}.

\begin{figure}[h!]
\centering
\includegraphics[width=\textwidth]{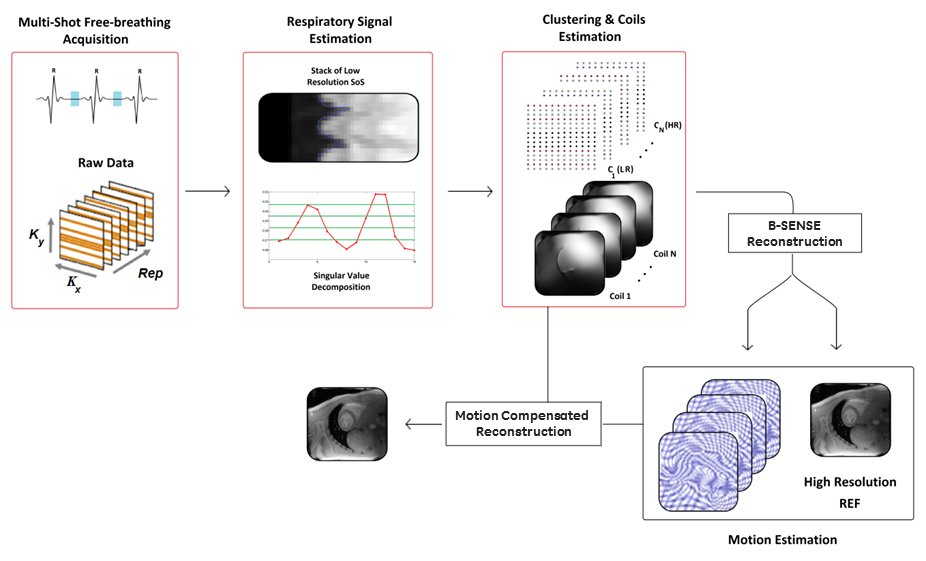}
\caption{Schematic illustration of the proposed reconstruction, including the non-rigid motion extraction. Acquisition is performed using complementary trajectories, leading to uniform samplings in the phase encoding direction, which allows for an optimal combination of the k-spaces according to their positions in the breathing signal. The motion model, initialized by registering the images from each respiratory bin, is incorporated into the reconstruction process.}
\label{fig:fig2}
\end{figure}

\section{Material and Methods}

The proposed reconstruction algorithm was applied and validated with different experiments using Matlab (The MathWorks, Natick, MA) on a PC with Intel Xeon 3.3 GHz CPU and 64GB ram. The experiments were performed on 3T MR750w and 1.5T MR450w systems (GE Healthcare, WI, USA).

\begin{table}
  \centering
  \caption{Parameters used for the different experiments. The acquisition matrix size was 192 x 256.}
  \label{tab:table1}
 \begin{tabular}{cccccccc}
    \toprule
     & \#repetition & \#calib & \#periphery & acc & acc shot & acc shot & NEX\\
     & $n_r$ & lines & lines & peri & pre-bin & post-bin & sequence\\
    \midrule
    Simulation 1 & 4 & 32 & 48 & 3.3 & 2.4 & - & 1.67\\
    Simulation 2 & 4 & 32 & 32 & 5 & 3 & - & 1.33\\
    Phantom      & 6 & 32 & 48 & 3.3 & 2.4 & - & 2.5\\
    In vivo      & 15 & 17 & 43 & 4.1 & 3.2 & 1.1 (5 bins) & 4.7\\
    \bottomrule
  \end{tabular}
\end{table}

\subsection{Offline Simulation on Synthetic Data}

In order to perform a realistic simulation, we first created a synthetic dataset based on actual LGE patient images. In one patient with suspected cardiovascular disease, four repetitions of a cardiac-gated, inversion recovery prepared, single-shot LGE scan were acquired in free-breathing 10 minutes after Gadolinium injection. Cardiac images were obtained with a spoiled fast gradient echo sequence and the following parameters: matrix size 192 x 256, in-plane spatial resolution 1.52 mm x 1.52 mm in short axis with slice thickness = 8 mm, readout flip angle = 20 degrees, echo time (TE) = 2.02 ms, mid-diastolic trigger delay, pulse repetition time (TR) = 4.43 ms and SENSE factor = 2 with partial Fourier. Synthetic k-space data were created by the application of synthetic coil sensitivity maps (with Gaussian profiles) to the LGE images, Fourier transformation and undersampling in the phase encoding direction. A full sampling of the central k-space area (17 lines) was used and the peripheral area was undersampled with a Golden Step Cartesian trajectory \cite{Derbyshire2011} with an acceleration factor $R = 3.3$. Spacing between samples was proportional to the Golden ratio ($p = 0.618$). This trajectory enables an irregular but almost uniform distribution of the acquired data for any arbitrary number of repetitions, leading to incoherent aliasing (Fig. \ref{fig:fig1}, right). The motion-free image was reconstructed using our reconstruction and compared to a motion correction method similar to that proposed by Kellman et al. \cite{Kellman2005}, where the motion-free image is recovered by averaging the registered images obtained after the B-SENSE reconstruction step. We call this prior method reconstruction-registration-average (RRA).

\subsection{Phantom Imaging}

Single-shot pulse sequences were used to acquire phantom images with a 26-channel cardiac coil. The sequence was modified to take into account the same Golden ratio sampling as in our offline simulation experiment. The protocol was applied to acquire phantom images with a resolution of 1.48 mm x 1.48 mm. A translational motion was imposed to the table to mimic respiratory motion.

\subsection{In Vivo Validation Experiment with Self-Navigation}

In vivo cardiac datasets from two healthy adult volunteers were acquired on a 1.5T scanner using a 32-channel cardiac coil. A multi-shot slice (15 shots) of a free-breathing, cardiac-gated, spoiled fast gradient echo sequence (without inversion-recovery preparation) was collected with the following parameters: TE = 2.10 ms, TR = 4.52 ms, 8 mm slice thickness, FOV = 253 mm x 338 mm, matrix size 192 x 256 and 1.32 mm x 1.32 mm in-plane resolution, diastolic trigger delay. A fully sampled reference was acquired additionally in breath-hold for visual comparison. Each shot consists of 60 k-space lines: the central k-space was fully sampled with 17 lines and the periphery (43 lines) was undersampled, leading to a global acceleration factor of 3.2. An estimate of the respiratory signal was extracted using the proposed self-navigated technique, and was subsequently used to separate the acquired data into five respiratory bins. An overview of the parameters used in this study is given in Table \ref{tab:table1}.

\section{Results}

The time needed to run the motion-compensated reconstruction for 15 shots of matrix size equal to 192 x 256 with 32-channel cardiac coils was about 1 min 35 s, including the time to compute the motion between shots.

\begin{figure}[h!]
\centering
\includegraphics[width=0.9\textwidth]{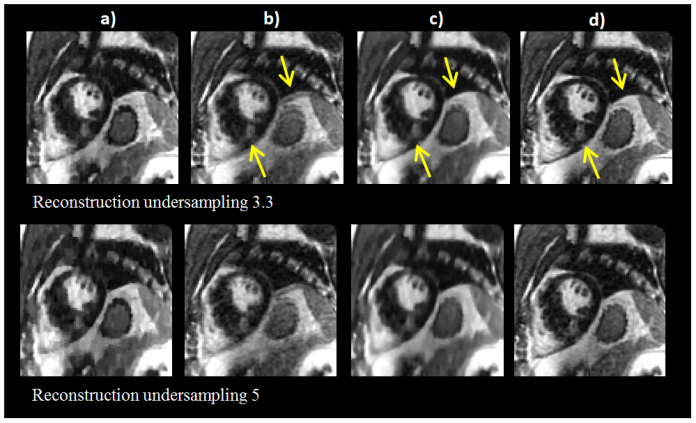}
\caption{Cardiac short-axis reconstruction of a synthetic dataset generated from 4 single-shot LGE acquisitions in free-breathing on a 37-year-old patient with acceleration factors $r = 3.3$ and $r = 5$. a) One reconstruction using a classic SENSE (192 x 256), b) Sum-of-Squares (all repetitions), c) Reconstruction-Registration-Average, d) proposed reconstruction.}
\label{fig:fig3}
\end{figure}

\subsection{Offline Simulation on Synthetic Data}

Example reconstruction results on the simulated data generated from a patient with nonischemic cardiomyopathy are shown in Fig. \ref{fig:fig3}. One can see a spatially blurred result with a standard reconstruction-registration-average (RRA) method. The proposed reconstruction exhibits significant quality improvements over each method with an acceleration factor $r = 3.3$ while reconstructing sharper edges (arrows) and small structures. For higher acceleration factors the performance of our method is much better compared to RRA, both in terms of reconstruction accuracy and image quality.

\subsection{Phantom Imaging}

\begin{figure}[h!]
\includegraphics[width=\textwidth]{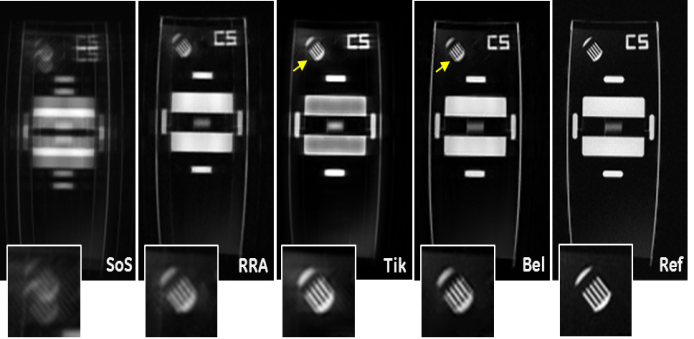}
\caption{Reconstructions on a phantom using two different regularization methods with acceleration factor 3.3. Six single-shot repetitions have been acquired. From left to right: Sum-of-Squares (SoS), Reconstruction-Registration-Average (RRA), Tikhonov, Beltrami, Reference.}
\label{fig:fig4}
\end{figure}

Similar results can be observed in phantom experiments (Fig. \ref{fig:fig4}). Comparisons with a classic Tikhonov reconstruction are shown in Fig. \ref{fig:fig4}. The results present the reconstructed phantom motion experiments where here the motion has been applied with the table. The sum-of-squares reconstruction (Fig. \ref{fig:fig4}, left) clearly exhibits the effect of motion. As in the previous experiment, the RRA method exhibits blurry results (due to the undersampling), although providing a motion-corrected denoised image. A visual improvement can be noticed when applying a motion compensated reconstruction with Tikhonov regularization. The latter method performs well but is, however, unable to recover sharp edges and some residual artifacts can still be seen on the recovered image. The use of a fast primal-dual algorithm combined with Beltrami regularization makes the proposed reconstruction robust with better performance in terms of image quality, reduced artifacts and sharpness (Fig. \ref{fig:fig4}, Bel).

\subsection{In Vivo Validation Experiment with Self-Navigation}

\begin{figure}[h!]
\centering
\includegraphics[width=0.6\textwidth]{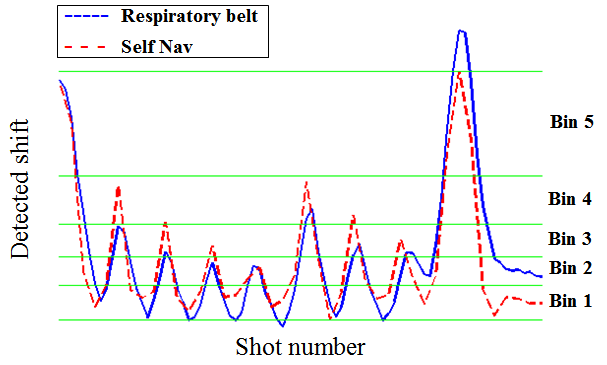}
\caption{Data binning step of the proposed self-navigated signal obtained from 150 consecutive 2D fast spoiled gradient echo acquisitions in free-breathing liver imaging.}
\label{fig:fig5}
\end{figure}

\begin{figure}[h!]
\centering
\includegraphics[width=\textwidth]{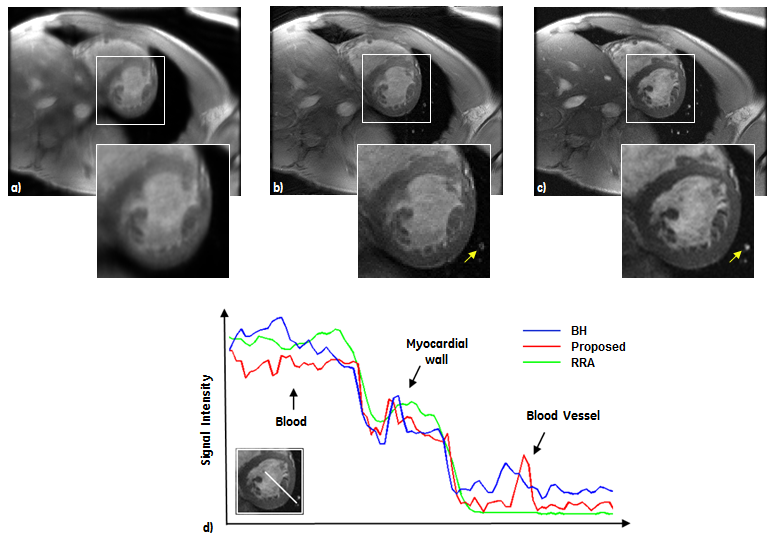}
\caption{Data binning step of the proposed self-navigated signal obtained from 150 consecutive 2D fast spoiled gradient echo acquisitions in free-breathing liver imaging.}
\label{fig:fig6}
\end{figure}

Figure \ref{fig:fig5} shows an example of the proposed fast and automatic self-navigated binning method on 150 consecutive slices of liver SPGR acquisition. The temporal rate was 400 ms, corresponding to a total acquisition duration of 1 min. The extracted respiratory signal (red) shows good agreement with the respiratory belt placed on the subject's thorax (blue), with a coefficient of determination $R^2 = 0.76$. Raw data acquired in similar motion states can be clustered into a reduced number of motion states, thereby improving the quality of images from which to extract motion in free-breathing without the need for navigators or external sensors.
Short-axis images of the myocardium of a healthy subject without Gadolinium injection and without inversion recovery preparation are shown in Fig. \ref{fig:fig6}. Both cardiac structures (myocardium wall, papillary muscles) and non-cardiac structures (blood vessels) are very well recovered with our reconstruction. The method yields significant sharpening of the myocardium wall and papillary muscles. However, due to the relatively high-undersampling, the RRA method is unable to recover a good quality image, exhibiting blurry structures and losing some of the details in the image such as blood vessels (arrows). This particularity is also seen in Fig. \ref{fig:fig6} d where a specific intensity profile is plotted. The sharpness of the edges on our motion-corrected reconstruction is confirmed as well as the fidelity to the breath-hold acquisition.

\section{Discussion and Conclusion}

We introduced a new free-breathing single-shot LGE pipeline, including an optimized sampling and the associated joint reconstruction and motion correction algorithm designed for fast and robust cardiac imaging. By incorporating the estimated motion into the reconstruction process, we increased the robustness of the model and exhibited good quality images.

In this study, we used a fast and automatic self-navigated binning strategy that aims to cluster the acquired raw data into similar motion states. While preliminary results have shown improved image quality and better motion estimation, additional optimization of number of bins and number of repetitions is still required to maintain an optimal tradeoff between reconstruction quality, reconstruction time and accuracy of motion estimates. The motion corrected images show better visual quality than classic reconstructions but appear less sharp than corresponding breath-held acquisitions, especially for high accelerations. Possible explanations are the inaccuracies in motion estimates or other effects related to MR physics, such as spin history or changes in $B_0$ and $B_1$ inhomogeneities induced by breathing.

One interesting application of the proposed motion correction model is for high-resolution 3D isotropic LGE single-shot imaging of the heart, such as the one proposed recently in \cite{Dzyubachyk2015} for myocardial scar assessment. This will allow for the reconstruction of 3D isotropic motion corrected volumes by keeping the advantages of a 2D acquisition (high tissue and vessel contrast, short acquisition time), e.g. using super-resolution techniques \cite{Odille2015}. Other applications, such as abdominal imaging \cite{Buerger2013} and coronary vessel imaging \cite{Cruz2016}, are being investigated.

A limitation to the method is that potential through-plane motion cannot be corrected, although it remains small compared to the slice thickness. To overcome this problem, one could consider weighting the images according to the motion amplitude compared to the target image or acquiring 3D slab instead of 2D slice data and applying motion compensation. The preliminary results presented in this work should be confirmed with further patient studies.

The feasibility of the proposed reconstruction has been evaluated in simulation, phantom and volunteer experiments. The method has been shown to allow non-rigid motion correction while efficiently recovering features, thanks to the Beltrami regularization scheme. The conventional segmented LGE acquisition is limited by the maximum breath-hold time, which limits the signal-to-noise ratio and/or spatial resolution. This limitation is overcome by the presented free breathing approach. Ultimately, this method could enable accurate motion corrected reconstruction of single-shot images with higher spatial resolution and a higher signal-to-noise ratio compared to conventional segmented methods, with the potential to offer high-quality LGE imaging in challenging patients.

\subsection*{Acknowledgments}
The authors thank Mayo Clinic (Rochester, MN), Advanced Cardiovascular Imaging (New York, NY) and Morriston Hospital (Swansea, UK) for providing some of the imaging data. This publication was supported by the European Commission, through Grant Number 605162. The content is solely the responsibility of the authors and does not necessarily represent the official views of the EU.

\bibliographystyle{splncs}
\bibliography{main_document}


\end{document}